\documentclass{article}
\usepackage{hyperref}
\usepackage{graphicx}
\usepackage{amsmath}
\usepackage{authblk}  
\usepackage{amssymb}
\usepackage{amsmath, amssymb, amsfonts}
\usepackage{url}
\usepackage{doi}

\newtheorem{note}{Note}

\title{Perception-Informed Neural Networks: Beyond Physics-Informed Neural Networks}
\author[1]{Mehran Mazandarani\thanks{Email: me.mazandarani@ieee.org}} 
\author[2]{Marzieh Najariyan\thanks{Email: marzieh.najariyan@gmail.com}}
\affil[1]{Senior Member, IEEE}
\affil[2]{Member, IEEE, Ferdowsi University of Mashhad}
\date{}  
\DeclareUnicodeCharacter{200B}{}

\begin{document}

\maketitle

\section{Abstract} 
This article introduces Perception-Informed Neural Networks (PrINNs), a framework designed to incorporate perception-based information into neural networks, addressing both systems with known and unknown physics laws or differential equations. PrINNs extend the concept of Physics-Informed Neural Networks (PINNs) and their variants, offering a platform for the integration of diverse forms of perception precisiation, including singular, probability distribution, possibility distribution, interval, and fuzzy graph. PrINNs allow neural networks to model dynamical systems by integrating expert knowledge and perception-based information through loss functions, enabling the creation of modern data-driven models. 
Some of the key contributions include Mixture of Experts Informed Neural Networks (MOEINNs), which combine heterogeneous expert knowledge into the network, and Transformed-Knowledge Informed Neural Networks (TKINNs), which facilitate the incorporation of meta-information for enhanced model performance. Additionally, Fuzzy-Informed Neural Networks (FINNs) as  a modern class of fuzzy deep neural networks leverage fuzzy logic constraints within a deep learning architecture, allowing online training without pre-training and eliminating the need for defuzzification. 
PrINNs represent a significant step forward in bridging the gap between traditional physics-based modeling and modern data-driven approaches, enabling neural networks to learn from both structured physics laws and flexible perception-based rules. This approach empowers neural networks to operate in uncertain environments, model complex systems, and discover new forms of differential equations, making PrINNs a powerful tool for advancing computational science and engineering.\\

\textbf{Keywords}:
\small Artificial Intelligence; Machine Learning; Computational Science and Engineering; Generalized Theory of Uncertainty;   \small Deep Neural Networks; Perception Computing; Dynamical Systems.

\section*{Introduction}

Humans have almost always been interested in acquiring knowledge about and perceiving the phenomena existing in the universe. To gain such knowledge, they have primarily made significant efforts to model these phenomena, mainly by utilizing the laws of physics\footnote{The laws of mechanics, electromagnetism, and thermodynamics.}. These laws have, in fact, originated from our perception\footnote{By perception, we mean observations, opinions, intuitiveness, experiments, experiences, and feelings.} of phenomena or systems, which have been precisely formulated in a mathematical form. As a result, models derived from these physics laws generally align with our perception\footnote{This alignment underlies the mode of precision.}. These models typically take the form of differential equations and are often referred to as physics-based models.   \\

 In recent decades, advancements in neural networks and machine learning have opened the door to presenting models using data, often vast amounts of data. Neural network-based models (NN-models) can be classified as data-driven models. Impressively successful applications of NN-models across various scientific fields demonstrate that they generally align with our perceptions, to an acceptable extent. This alignment is largely due to the influence of the vast amounts of data fed into these models. In fact, NN-models are informed of our perceptions solely through numerical data, which consists of measurement-based information. However, numerical data often lacks the capacity to fully represent perceptions. \\
  
 This raises an important point: Whether there is another way to inform NN-models of our perceptions. In other words, whether NN-models can be guided toward satisfying our perception, at least in certain regions corresponding to the system. Another question is whether the output of NN-models can satisfy our perception, even when the data fed into the models is less than what is typically required. The results of examining NN-models with limited data have shown that these models may fail to align with our expectations, even to some extent.\\

The concept of physics-informed neural networks (PINNs) \cite{PINN-1, PINN-2, PINN-3} can be considered a significant step toward addressing the questions mentioned above. Broadly speaking, PINN models can be defined as those in which the neural network is not only fed with data, often much less than NN-models, but also with physics-based models, or, more generally, with physics laws.
In essence, PINNs ensure that the model is aware of what has already been established through an agreement with our perception. This enforcement is achieved by incorporating the corresponding physics-based model\footnote{Physics-based model with initial and boundary conditions, or, in a general case, physics laws. See \cite{PINN-rev} for more details.}, typically expressed as partial differential equations (PDEs), into the total loss function as an additional term.  \\

The successes of PINN models, as demonstrated in \cite{PINN-rev}, highlight the effectiveness of this approach, in which information about the dynamical system, represented by differential equations, is shared with NN-models. In fact, the applicability of PINNs can be viewed from several perspectives, two of which are briefly explained as follows. First, in cases where data for the system is limited and the differential equations describing the system are known, the solution to these equations may not be readily available, and conventional neural networks may fail to model the system with limited data. In such instances, PINNs, by integrating differential equations and the limited data fed into the neural networks, provide a pathway to address the issue, as discussed in \cite{PINN-1}.  \\

In the second case, there is some data available for the system, and the general structure of the differential equations governing the system is known\footnote{In general, the structure of the differential equations governing the system may be partially known.}. However, some parameters within this structure are unknown. In this situation, PINN models, using known numerical data and the provided differential equations, can approximate the unknown parameters while simultaneously learning the system model.  \\

In the third case, there may be limited or an apparently acceptable amount of data for the system, along with the physics laws governing it. However, no sufficient information about the corresponding differential equations is available. In such a case, PINNs compel the network to learn not only from the data but also from the physics laws governing the system. This approach makes it almost possible to have a system model that aligns with our perception, at least to some extent. \\

As a result, PINNs rely on integrating differential equations, more specifically, crisp differential equations, or, in a general case, physics laws, which serve as a mode of precisiation of perceptions. However, there are many other modes for precisiating perceptions, including interval values, probability distributions, possibility distributions, fuzzy graphs, bimodal distributions, and so on. In fact, the mode of precisiation of perception in PINNs is singular, meaning that the information about the initial conditions, boundary conditions, and relationships between the states or variables of the system, expressed in the form of differential equations, is considered certain or crisp. Nevertheless, uncertainty is an inherent attribute of perception and, indeed, of information. Therefore, uncertainty in data is inevitable. In PINNs, the network is enforced to adjust itself or to attempt not to be misled by the provided system data, which is inherently uncertain, while also considering a description of the system,  crisp differential equations, that is certain. There are a few comments regarding models based on neural networks or, more generally, machine learning methods, in which information about dynamical systems is integrated.\\

First, uncertainty is an intrinsic attribute of dynamical systems and plays a key role in developing models intended to reflect reality more accurately. For instance, consider the well-known relationship $ (F = mg) $. Although $ (g) $ is often treated as a constant, in practice, its value is uncertain; it may vary slightly from one location to another due to differences in altitude, latitude, or local geological structures. However, in many models, this constant is assumed to be known and fixed. Such an assumption about the system's parameters, initial conditions, and boundary conditions leads to a simplified model that might not fully capture the inherent variability of the actual system. Additionally, there are different types of uncertainty, and one or more of these may be considered depending on the application. Recognizing and incorporating uncertainty in parameters, initial conditions, and boundary conditions of different types can result in models that more faithfully represent the complexities of real-world behavior. \\

Second, many systems or phenomena do not neatly align with classical physics laws or lend themselves to formulation through differential equations. This is especially true for domains such as economics, social dynamics, and certain biological systems, where human behavior, complex interactions, and emergent properties dominate. As a simple example, consider the dynamics of the stock market, where the interplay of investor sentiment, unpredictable external events, and feedback loops often defies a straightforward description using standard differential equations. Nevertheless, there might be some information about the system's dynamics that can be shared with the neural network to enhance the efficiency of the resulting model.    \\

Third, in real-world systems, the behavior of dynamical systems is often too complex to be captured perfectly by crisp differential equations or, more generally, by differential equations. To make the modeling process manageable, many influencing factors and uncertainties are deliberately simplified or even ignored. For example, consider a swinging pendulum in an engine. In reality, the pendulum's motion is affected by factors such as friction, air resistance, and slight imperfections in the pivot. Moreover, there may be additional influences that we are not even aware of, making it impossible to include every relevant factor in the model in terms of differential equations. In other words, there are cases in which modeling the system using only differential equations may not always be sufficient to approximate the complexities of the actual dynamical system to an acceptable extent. In such cases, transferring knowledge about the system into the neural network may extend the model's capacity to better align with reality.  \\
Fourth, there are many cases of dynamical systems where experts' perceptions and precisiations of the parameters, initial conditions, boundary conditions, or the local dynamics may not align with each other. As a simple example, consider the factor $ k $ in $ \dot{x}(t) = -kx(t) $, for which the perception is that it is about 2. One expert may precisiate the value of $ k $ as the interval number $ k=[1.5, 2.5] $, while another expert may treat $ k $ as a random variable with a probability distribution. In such cases, the model's performance may be improved by integrating different experts' perception precisiation.\\

The thesis of perception-informed neural networks (PrINNs) introduced in this paper builds upon the insights presented in the above comments. Moreover, it extends beyond these initial observations, representing a significant step toward enabling neural networks, and more broadly machine learning methods, to directly incorporate perceptions about a system through words, propositions, and imprecise differential equations (IDEs) \cite{IDE}. Indeed, PrINNs allow for dynamic adjustments to model parameters based on perceived data. What is crucial is that PrINNs do not reject PINNs but rather embrace them as one mode alongside others. In fact, PrINNs establish a framework for integrating perceptions with diverse precisiation into neural networks, thereby enabling the creation of models that incorporate expert knowledge about the system. Furthermore, PrINNs pave the way for transferring this knowledge to neural networks, empowering them to improve their comprehension of complex information through perception-based rules, particularly when no established physics laws or differential equations are available. Finally, PrINNs offer a valuable platform for exploring previously unknown forms of differential equations, modern data-driven models, and a deep insight into the pivotal role of perception and IDEs in constructing data-driven models based on neural networks.\\

This article aims to outline PrINNs as a conceptual platform designed to explore the origin of informed neural networks presented so far, as well as the opportunities and methodologies for devising and introducing new informed neural networks. It also emphasizes the crucial role of perception in this context. Rather than focusing on technical implementation details, the article concentrates on explaining the theoretical foundations and potential pathways for the development of these models.\\

The core of PrINNs is tied to perception-based information. To utilize such information, PrINNs require a precise meaning of it, which is achieved through the precisiation process underlying the generalized theory of uncertainty \cite{gtu-1, gtu-2}. This process is discussed in the context of dynamical systems in the next section.

\section{The Precisiation of Perception-based Information in Dynamical Systems}

One remarkable human capability is our ability to tolerate imprecision and uncertainty in perception. For instance, you might perceive the weather as \textit{hot} based on your overall impression. When asked to precisiate \textit{hot}, you might respond with approximately 38°C\footnote{By approximately 38°C, you may mean an interval, a probability distribution, a fuzzy number, etc.}, even though the thermometer reads 37.5°C. The next day, even if the temperature slightly increases to 38.5°C, your overall impression remains \textit{hot}. This demonstrates that the thermometer reading aligns with your perception. In fact, our perception is capable of handling and computing with imprecise information received from our sensory systems and cognitive processes. One of the main goals in perception-informed neural networks is to enable neural networks to possess such a capability. \\

PrINNs can be defined as neural networks informed by our perception of a dynamical system, described through words or, more generally, propositions drawn from natural language. Words and propositions convey information, which is often expressed in imprecise forms. Therefore, to inform the network of our perception, it is necessary to precisiate perception-based information. Thus, we first need to understand how to achieve such a precisiation, which serves as an important starting point for introducing various facets of PrINNs. What follows briefly elaborates on some concepts drawn from the computational theory of perception (CTP), the generalized theory of uncertainty (GTU), precisiated natural language, and IDEs in the context of dynamical systems. For more details, see \cite{gtu-1, IDE, ctp}. \ \\

\begin{figure}[ht!] 
	\centering
	\includegraphics[scale=0.22]{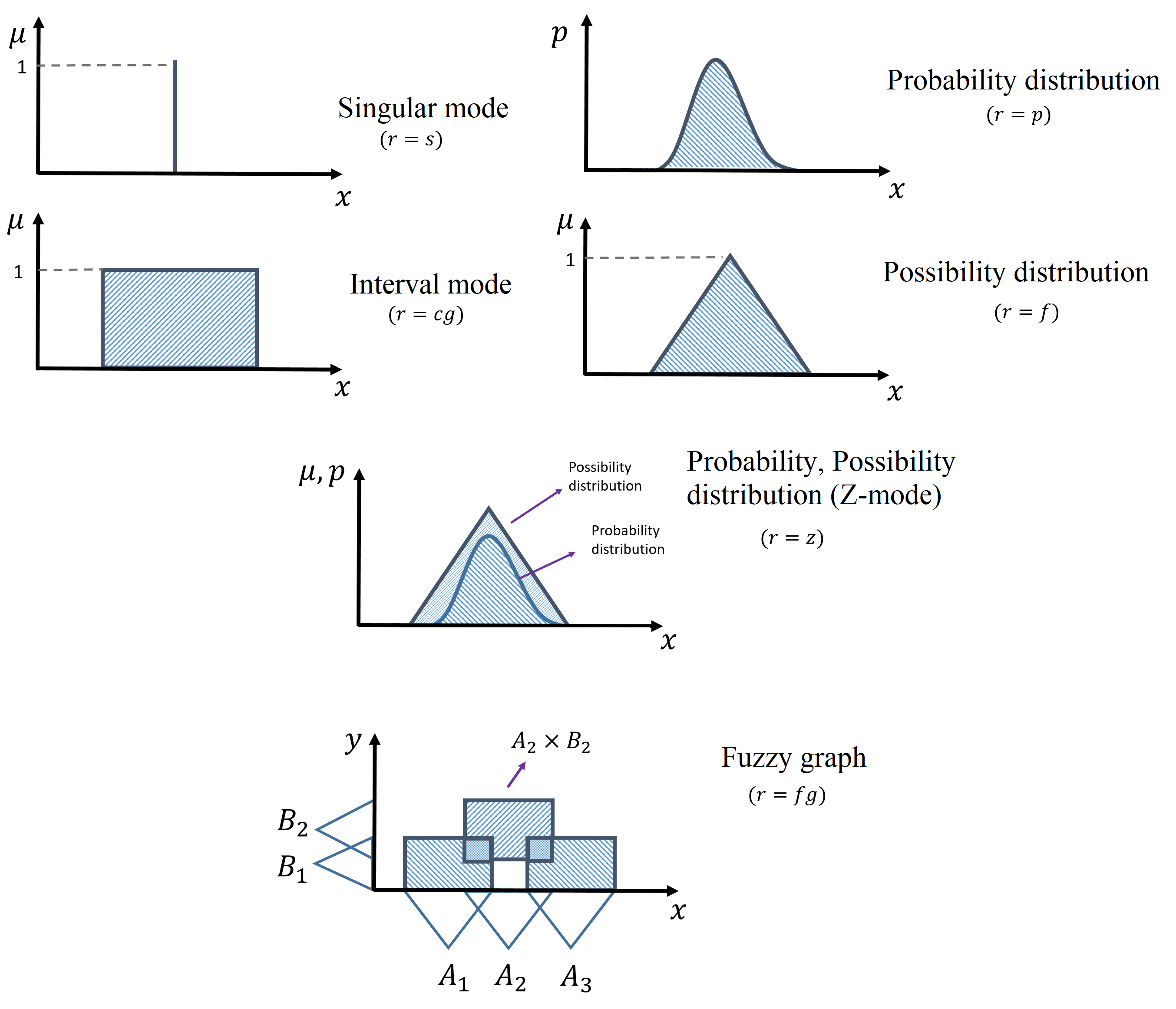}
    \caption{Some of the modes of precisiation in generalized theory of uncertainty}
    \label{fig1}
\end{figure}

Perception encompasses a broad range of human cognitive and sensory interpretations that are inherently imprecise, fuzzy, and context-dependent. Moreover, perception includes elements such as opinions, experiences, observations, experiments, intuition, and feelings. Perception is often expressed in natural language through words or propositions, serving as a correspondent that conveys imprecise information. Simple examples in this regard include:
\begin{itemize}
\item The variation of the variable $ x $ with respect to $ t $, is \textit{high}, when $ t $ is \textit{about 3}
\item The pressure is \textit{low} at $ t=5 $
\item The initial condition is \textit{near zero}
\item If customer satisfaction is \textit{low}, then purchase frequency is \textit{low}
\item The temperature distribution is \textit{close to $100^\circ $}C when the spatial location is \textit{near the boundary point L}
\end{itemize}
In the propositions above, the italicized words are imprecise, and a key question that arises is how to interpret such propositions. Precisiation is the process of transforming imprecise propositions into a structure called a generalized constraint (GC). The generalized constraint of a proposition, pp, has the following formalized framework:
\begin{equation*}
GC(p): \ X \ \ \text{isr} \ \ R
\end{equation*}

where $ X $ is the focal variable, parameter, or state of the system constrained by $ R $. For instance, $ X $ can represent pressure, temperature, angle, velocity, satisfaction, price, and so on. Additionally, rr defines the mode of precisiation, with important modes including possibility distribution (r = $ f $), probability distribution (r = $ p $), interval or crisp granulation (r=$ cg  $), fuzzy graph (r = $ fg $), singular (r=$ s $), and when undetermined, $ r $ is blank\footnote{In GTU terminology \cite{gtu-1}, r = \text{blank} means possibility distribution. However, in this article, it is interpreted as an undetermined mode.}. Fig. \ref{fig1}, adopted from \cite{gtu-1,gtu-2}, illustrates several modes of precisiation. Moreover, $ R $ is a constraining relation; for example, $ R $ could be a normal distribution function.
As an example, the GC form of $p$: "The variation of the variable $ x $ with respect to $ t $ is \textit{high}, when $ t $ is \textit{about 3}" in the mode of probability distribution is:
\begin{equation*}
GC(p): \ \dot{x}(t)|_{t \ isp \ \mathcal{N}(3,0.1)} \ \ \text{isp} \ \ \mathcal{N}(5,1)
\end{equation*}
 where $ \mathcal{N}(m,\sigma^2) $ denotes the normal distribution function with mean $ m $ and variance $ \sigma^2 $.  \\
 
Since in PrINNs the network is designed to deal with dynamical systems, what follows extends the precisiation of perception to dynamical systems from two perspectives: (1) There is an imprecise description of the dynamical system in the form of differential equations, and (2) The description of the dynamical system or parts of its dynamics are expressed by propositions.

\subsection{Dynamical Systems with Known Differential Equations} 
For the sake of simplicity, consider a dynamical system \( S \) whose description has been mathematically precisiated by the following IDE:
\begin{equation} \label{eq1}
S^*: \begin{cases}
\dot{x}(t) = f(t, x; \lambda), \\
x(t_0) = x_0,
\end{cases}
\end{equation}
where $ S^* $ denotes a mathematical precisiation of $ S $, \( x \) represents the state of the system, i.e. the solution of the differential equation, \( x_0 \) is the initial condition, and \( f \) is a nonlinear function parameterized by \( \lambda \). {Moreover, the initial condition, $ x_0 $, or a subset of parameters, $ \lambda $, is described by propositions conveying perception-based information about their values.}.  A simple instance of (\ref{eq1}) may be presented by 
\begin{equation} \label{eq2}
S_1^*: \begin{cases}
\dot{x}(t) = \lambda x(t), \\
x(t_0) = x_0, \\
\lambda  \text{ is small and negative}, \\
x_0 \text{ is approximately 5},
\end{cases}
\end{equation}

As observed, the IDE (\ref{eq2}), extrinsically, i.e., in terms of surface structure and the resulting system dynamics, provides a precise mathematical description of $ S_1 $. For illustration, the expression $ \dot{x}(t) = (\text{small and negative})\cdot x(t) $ implies that the variation of $ x $ with respect to time is small and negatively proportional to its current value. However, since perception-based information is intrinsically imprecise, an IDE also inherently represents an imprecise description of $ S $ in an abstracted form. Thus, an IDE may be interpreted as an imprecise mathematical precisiation (im-precisiation), or more specifically, as a machine-oriented precisiation (m-precisiation) of $ S $. Hence, the m-precisiation of $ S $ is denoted by $ S^* $.\\

\begin{figure}[ht!] 
	\centering
	\hspace*{-1cm}
	\includegraphics[scale=0.28]{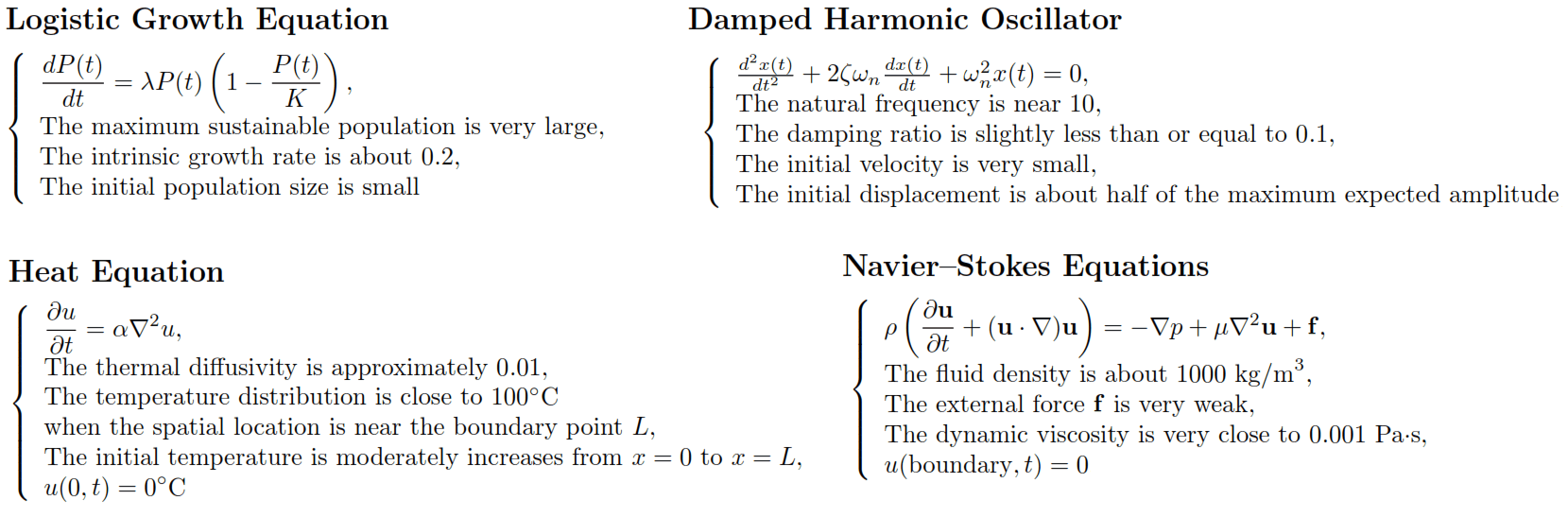}
    \caption{Some instances of IDEs.}
    \label{fig2}
\end{figure}

The differential equations (\ref{eq1}) and (\ref{eq2}) belong to a class of IDEs called ordinary IDEs (OIDEs). OIDEs are ordinary differential equations in which some parameters, initial conditions, or boundary conditions are characterized generally by propositions. Similarly, partial IDEs (PIDEs) are defined as partial differential equations that include linguistic variables, words, or propositions.
There are also other types of IDEs, such as fractional IDEs, integro-IDEs, and so on, which can be defined in a similar manner to OIDEs and PIDEs. In fact, an IDE embodies a combination of our knowledge from physics laws and perception-based information \cite{IDE}. Additionally, IDEs underpin a general form of calculus, which may be called perception-based calculus. This is, in essence, the calculus of impression, in which computing with perception plays a pivotal role. 
As an illustration, Fig.~\ref{fig2} provides examples of some IDEs. \\

Thus, roughly speaking, IDEs are the result of integrating perception into differential equations, for which determining a solution requires a precise meaning of the propositions that convey perception. The process of precisiation of an IDE involves the transformation of perception-based information expressed about the differential equation entities. Therefore, any initial conditions, boundary conditions, or parameters described by words or propositions must be transformed into their GC form with a specific mode.

As a simple example, consider the system $S_1$ whose corresponding IDE is given in (\ref{eq2}). The GC of the IDE thus corresponds to the GC of $ S_1^* $. If the mode of precisiation is considered singular, it implies that $ \lambda $ and the initial condition, $x_0$, are crisp values. For instance, they may be characterized by $ \lambda = -0.5 $\footnote{Characterizing the meaning of "small and negative" for $\lambda$ depends on our perception or expert knowledge.} and $x_0 = 5$. Therefore, the precisiation of IDE (\ref{eq2}), or the nested precisiation of system $S_1$, in singular mode, is denoted by $GC (S_1^*)|_{r=s}$, as follows:

\begin{equation} \label{eq3}
GC (S_1^*)|_{r=s}: \begin{cases}
\dot{x}(t) = -0.5 x(t), \\
x(t_0) = 5, 
\end{cases}
\end{equation}
As seen, the singular precisiation of IDE (\ref{eq2}) has resulted in the crisp differential equation (\ref{eq3}). More generally, symbolically, we can write
 $$GC (IDEs)|_{r=s} =\text{crisp differential equations}$$
  
\begin{figure}[ht!] 
	\centering
	\hspace*{-1cm}
	\includegraphics[scale=0.22]{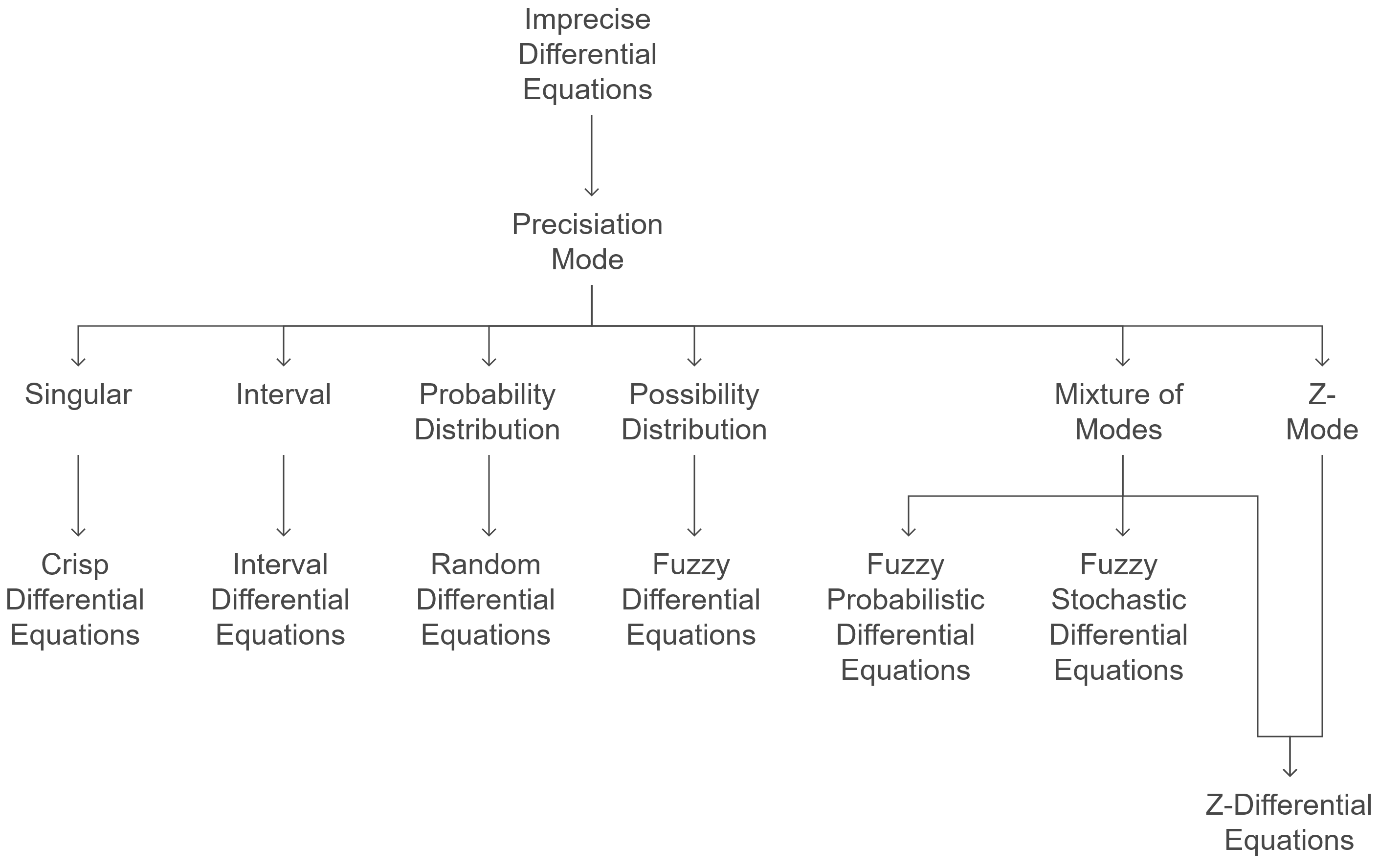}
    \caption{The precisiation of IDEs in different modes.}
    \label{fig3}
\end{figure}

  Therefore, depending on the mode of precisiation, a system described by IDEs can be transformed into a system described by interval differential equations, random differential equations, fuzzy differential equations, etc., provided that the mode of precisiation is interval (i.e., r=$ cg $), probability distribution (i.e., r=$ p $), possibility distribution (r=$ f $), and so on, as shown in Fig. \ref{fig3}. Hence, except in the case of the singular mode, IDEs, when precisiated, generally result in uncertain differential equations (UDEs). Symbolically, we can write:
  $$ GC(IDEs) = UDEs $$

As a matter of fact, IDEs provide us with a formal structure of differential equations in which the relationships between the states of the system, as well as the initial and boundary conditions, can be described by propositions conveying perception-based information. Therefore, the solution of an IDE is, in essence, a set of propositions, i.e., perceptions, as shown in Fig. 4 in \cite{IDE}. Nevertheless, at this juncture, there is no direct method to obtain the solution of an IDE. However, PrINNs can assist us in partially and approximately solving IDEs by utilizing the concept of precisiated modal solutions.

By definition, the precisiated modal solution of an IDE is the solution of the precisiated IDE in a specific mode. For instance, the solution to differential equation (\ref{eq3}) is a precisiated modal solution of the IDE (\ref{eq2}).

\begin{note} It should be noted that the modal solution of an IDE takes the form of propositions and represents perception-based information. However, the precisiated modal solution of an IDE represents measurement-based information. 
\end{note}

So far, we have discussed the precisiation of perception in IDEs by considering only a single mode. Precisiation in the single mode is associated with cases in which an expert, or a group of experts, associates the impression in the entire IDE with a fixed and uniform mode of precisiation. As a result, r is indexed by a single mode in the GC form, $ X $ isr $ R $.

However, experts, or more generally people, may have differing views on the mode of perception precisiation. In such cases, which frequently occur, the precisiation of an IDE is based on a mixture of experts (MoE) with differing opinions. This leads to a differential equation, or a set of differential equations, derived from a mixture of modes (MoM) of perception precisiation. Such differential equations may be called MoE-based differential equations.
In the following, we will demonstrate how an MoM precisiation of an IDE results in either an MoE-based differential equation or a set of MoE-based differential equations. Let us consider the following IDE as a mathematical model of a damped harmonic oscillator:

\begin{equation} \label{eq:damped_oscillator}
S^*:\left\{
\begin{array}{l}
\ddot{x}(t) + 2 \zeta \omega_n \dot{x}(t) + \omega_n^2 x(t) = 0,\\
\text{The natural frequency is near } 10, \\
\text{The damping ratio is approximately } 0.1, \\
\text{The velocity is small}, \\
\text{The initial displacement is about half of the maximum expected amplitude},
\end{array}
\right.
\end{equation}

For the sake of simplicity, assume that there are only two experts\footnote{There may also be just one expert who considers the different modes of precisiation.}, denoted by $ E_1 $ and $ E_2 $. Moreover, there is a consensus among the experts that the propositions pertaining to the velocity and initial displacement are interpreted as being equal to zero and two, respectively, i.e., they have been precisiated in the singular mode. However, expert $ E_1 $ associates the meaning of "the natural frequency is near $ 10 $" and "the damping ratio is slightly less than or equal to $ 0.1 $" with random variables that follow specific probability distributions. On the other hand, expert $ E_2 $ believes that these quantities are better represented as possibilistic, and thus a possibility distribution should be used to precisiate them. From this perspective, according to expert $ E_1 $'s opinion, IDE (\ref{eq:damped_oscillator}) is precisiated into a random differential equation (RDE), exemplified by:

\begin{equation} \label{eq:damped_oscillator-E1}
GC(S^*)|_{r=p,s}\left\{
\begin{array}{l}
\ddot{x}(t) + 2 \zeta \omega_n \dot{x}(t) + \omega_n^2 x(t) = 0,\\
\omega = \mathcal{N}(10, 0.1), \\
\zeta = \mathcal{N}(0.1, 0.01), \\
\dot{x}(t_0)=0, \ \ x(t_0)=2,
\end{array}
\right. 
\end{equation}

where $ GC(S^*)|_{r=p,s} $ means that the precisiation of system $ S $ (\ref{eq:damped_oscillator}) is based on two modes: the singular mode and the probability distribution mode. Moreover, $ \mathcal{N}(m, \sigma^2) $ denotes a normal distribution function with mean $ m $ and variance $ \sigma^2 $. However, if expert $ E_2 $'s opinion is considered for precisiating IDE (\ref{eq:damped_oscillator}), then the result of the precisiation is a fuzzy differential equation (FDE), exemplified by:

\begin{equation} \label{eq:damped_oscillator-E2}
GC(S^*)|_{r=f,s}\left\{
\begin{array}{l}
\ddot{x}(t) + 2 \zeta \omega_n \dot{x}(t) + \omega_n^2 x(t) = 0,\\
\omega = (9.8, 10, 10.2), \\
\zeta = (0.08, 0.1, 1.01), \\
\dot{x}(t_0)=0, \ \ x(t_0)=2,
\end{array}
\right.
\end{equation}

where $\omega$ and  $\zeta$ are triangular fuzzy numbers representing possibility distribution functions, denoted by the ordered triple $(a, b, c)$ with $a \leq b \leq c$. As seen, in this case, the effect of MoE on the precisiation of IDE (\ref{eq:damped_oscillator}) has resulted in a set of differential equations, namely RDE (\ref{eq:damped_oscillator-E1}) and FDE (\ref{eq:damped_oscillator-E2}), which together constitute MoE-based differential equations. \\

Indeed, the combination of differential equations (\ref{eq:damped_oscillator-E1}) and (\ref{eq:damped_oscillator-E2}) gives rise to what is called Z-differential equations (ZDEs) \cite{ZDE}, which may be expressed as bimodal differential equations combining FDEs and RDEs. Therefore, in this particular and yet important case, the precisiation of IDE (\ref{eq:damped_oscillator}) can be expressed in a compressed form as:

\begin{equation} \label{eq:damped_oscillator-E1E2}
GC(S^*)|_{r=z,s}:\left\{
\begin{array}{l}
\ddot{x}(t) + 2 \zeta \omega_n \dot{x}(t) + \omega_n^2 x(t) = 0,\\
\omega = ((9.8, 10, 10.2), \mathcal{N}(10, 0.1)), \\
\zeta = ((0.08, 0.1, 1.01), \mathcal{N}(0.1, 0.01)), \\
\dot{x}(t_0)=0, \ \ x(t_0)=2,
\end{array}
\right.
\end{equation}

where $\omega$ and $\zeta$ are $\mathbb{Z}^{+}$-numbers, denoted by the ordered pair $(\tilde{A}, p_x)$, where $\tilde{A}$ is a fuzzy number representing the possibility distribution of values that a real-valued uncertain variable $x$ can take, and $p_x$ is the probability density function of xx. Therefore, ZDEs can be considered a form of MoE-based differential equations. Additionally, ZDEs enable the integration of the concept of sureness—a fusion of probability and possibility distributions—into neural networks. This approach informs the neural networks of the logical principle that impossibility implies improbability, reflecting a consensus between two perspectives on perception precisiation. Such an approach results in models called sureness-informed neural networks (SINNets). In fact, SINNets emphasize the measure of sureness in their output. Furthermore, they underpin transformed-knowledge informed neural networks (TKINNs), which will be discussed in more detail in the next section.\\

In the second case, let us assume that both experts interpret the statement "\textit{natural frequency is near $10$}" as a random variable with a probability distribution function, exemplified by $\omega = \mathcal{N}(10, 0.1)$. However, they precisiate the perception of the damping ratio using a possibility distribution function, represented by the triangular fuzzy number $\zeta = (0.08, 0.1, 1.01)$. Consequently, affected by the MoE, the IDE (\ref{eq:damped_oscillator}) is precisiated as:

\begin{equation} \label{eq:damped_oscillator-E1E2-second case}
GC(S^*)|_{r=f,p,s}:\left\{
\begin{array}{l}
\ddot{x}(t) + 2 \zeta \omega_n \dot{x}(t) + \omega_n^2 x(t) = 0,\\
\omega = \mathcal{N}(10, 0.1), \\[4pt]
\zeta = (0.08, 0.1, 1.01), \\[4pt]
\dot{x}(t_0)=0, \ \ x(t_0)=2,
\end{array}
\right.
\end{equation}


As observed, the MoM precisiation of the IDE, affected by MoE in this case, results in a single MoE-based differential equation referred to as a \textit{fuzzy–probabilistic differential equation (FPDE)}\footnote{In some literature, FPDEs are termed fuzzy–stochastic differential equations; however, no stochastic process is involved in this context.}. In essence, FPDEs incorporate both probability and possibility distributions as the major modes of IDE precisiation, primarily affecting parameters, initial conditions, and boundary conditions. The significance of FPDEs lies in the fact that they can be employed to inform the neural network of a hybrid uncertainty, arising from a mixture of expert knowledge in the precisiation of the IDE. This enables the neural network to simultaneously account for both randomness and vagueness. Such neural networks, which may be called fuzzy–probabilistic informed neural networks (FPINNets), form a prominent class of PrINNs known as mixture of experts informed neural networks (MOEINNs). \\

Briefly, MOEINNs are neural networks in which a mixture of modes of precisiation of perception, primarily influenced by a mixture of experts, is used to precisiate IDEs or, more generally, the dynamics of the system described by propositions. Furthermore, it is the MoE-based differential equations that allow MOEINNs to be informed and enriched by a wide variety of expert knowledge regarding the system dynamics. MOEINNs, along with some of their subsets, such as TKINNs and SINNets, will be discussed in the next section.
%

\subsection{Dynamical Systems with Unknown Differential Equations} \label{Section2-Subsection-2}

Unlike the preceding section, where the differential equations of dynamical systems are known, this section deals with the precisiation of dynamical systems for which there are no known differential equations, or for which there is a simplified differential equation. Such systems are often found in, but are not limited to, non-mechanistic systems such as economic systems, biological systems, and social systems. The premise is that, in any case, whether or not the differential equations are known, parts of the system dynamics, some of the system states, or more generally the entire system, can still be described perceptually or by propositions. In other words, just as neural networks are informed of the physics laws governing the system (e.g., conservation laws), this section considers cases where the aim is to inform the neural networks of perception-based rules governing the system. The main difference between physics laws and perception-based rules is that the former are rigid, while the latter are generally elastic.  \\

Additionally, it should be noted that perception-based rules often provide a local and coarse description of the system, offering the network some flexible constraints. For simplicity, in this article, perception-based rules are presented in the form of fuzzy if-then rules\footnote{By fuzzy if-then rules, the description and precisiation of the system occur concurrently.}. In other words, the precisiation of dynamical systems is carried out in the fuzzy graph mode, i.e., $r=fg$. It should be noted that utilizing the fuzzy graph mode to precisiate a dynamical system is a well-known approach in the theory of fuzzy sets. What follows recalls the fuzzy graph mode of precisiation, adopted from \cite{gtu-1, gtu-2}, which will be used in the next section. \\

In the fuzzy graph mode, a dynamical system $S$, or some of its states, is precisiated by a disjunction of Cartesian products of fuzzy sets, represented as:

\begin{equation} \label{FG-mode1-1}
GC(S)|_{r=fg}: \ S \ \text{isfg} \ \sum_i^n A_i\times B_i 
\end{equation}
where $ \times $ denotes the Cartesian product, $ \sum_i^n $ represents the disjunction over multiple fuzzy set pairs, $ (A_i, B_i) $. Here, each Cartesian product $ A_i \times B_i $​ is characterized by the joint membership function, which is defined as:
\begin{equation}
 \sum_i^n A_i\times B_i \quad \text{means} \quad \sum_i^n \mu_{A_i} (x) \wedge \mu_{B_i}(y),
\end{equation}

where the operator $ \wedge $  represents a t-norm, $ \mu_{A_i} $ and $ \mu_{B_i} $ denote the membership functions of $ A_i $ and $ B_i $ in order. Indeed, the fuzzy graph in (\ref{FG-mode1-1}) is a representation of the following collection of fuzzy if-then rules 

\begin{equation} \label{FG-mode1-2}
R: \text{if } x \ \ \text{is} \ \ A_i, \ \ \text{then} \ \ y \ \ \text{is} \ \ B_i, \ \ i=1, 2, ..., n
\end{equation}
where $ x $ and $ y $ are the focal variables of the system. Additionally, by (\ref{FG-mode1-2}) $ S $ can be precisiated equivalently as $ GC(S)|_{r=fg}: S \ \text{isfg} \ R  $ where $ R $ is a restriction imposed on the joint constraint of $ x $ and $ y $.

Although there is a wide variety of cases that can be described by perception-based rules, a few of them are presented below.

 \textbf{Case 1}. Let us suppose that  $ x(t) $ denotes a state of $S$ and the local dynamics of $ x(t) $ with respect to time $ t $  are governed by the following rules  
 \begin{enumerate}
 \item[$ R_1 $]: if $t$ is small, then $ x(t) $ is large
 \item[$ R_2 $]: if $t$ is large, then $ x(t) $ is medium
 \end{enumerate}
 
The system $S$ may be considered a model of customer behavior in a retail environment, where $ x(t) $ represents the number of customers making a purchase at time $t$. The perceptual rules outlined above reflect an experiential understanding of customer behavior—observations accumulated over time. In this case, the first rule suggests that when the waiting time $t$ is short, the number of customers making a purchase, i.e., $ x(t) $, is perceived to be large. The second rule indicates that when the waiting time is long, customers tend to exhibit hesitation. In this scenario, even though some customers may still decide to make a purchase, the overall number of transactions $ x(t) $ is perceived to be medium. This is a simple example of a dynamical system where it may not be possible to determine any differential equations or physics laws to inform neural networks. Nevertheless, along with the data fed into the neural network, PrINNs can be informed by the perception-based rules governing the system. \\

 \textbf{Case 2}. Let $S$ be a dynamical system such that one of its states is denoted by $ x(t) $,  and the dynamical variations of $ x(t) $ with respect to time $ t $  are locally described as 
 
 \textit{The variation of $x(t)$ with respect to time $ t $ is small when $ t $ is near zero, and it is   approximately 2 when $ t $ is about 10}. 
 
 Such a description can be represented mathematically as 
 \begin{equation} \label{FG-mode1-3} 
 \begin{array}{l}
  \frac{dx(t)}{dt} |_{t = \text{near zero}} = \text{small}\\
  \frac{dx(t)}{dt} |_{t = \text{about 10}} = \text{about 2}
 \end{array}
  \end{equation}
In fact, the above instances illustrate a class of derivatives that may be referred to as \textit{perception-based derivatives}\footnote{Here, the perception-based derivatives have been precisiated in the mode of fuzzy graph.}. Alternatively, (\ref{FG-mode1-3}) can be represented in the form of fuzzy if-then rules as:
 \begin{enumerate}
 \item[$ R_1 $]: if $t$ is near zero, then $ \dot{x}(t) $ is small
 \item[$ R_2 $]: if $t$ is about 10, then $ \dot{x}(t) $ is approximately 2
 \end{enumerate}
 Therefore, the overall restriction imposed on the joint constraint of $ t $ and $ \frac{dx(t)}{dt} $, i.e. the precisiation of perception rules regarding the variation of $x(t)$ with respect to time $ t $, can be mathematically expressed as:
 \begin{equation} \label{FG-mode1-4}
R: \mu_{\text{near zero}}(t) \wedge \mu_{\text{small}}(\dot{x}(t)) + \mu_{\text{about 10}}(t) \wedge \mu_{\text{approximately 2}}(\dot{x}(t))
\end{equation}
It should be noted that although, in this case, the perception-based derivative has been illustrated with respect to time $ t $, the concept can be extended to partial perception-based derivatives, fractional perception-based derivatives, and beyond. Moreover, the precisiation of perception-based derivatives may be subject to other modes, such as probability distribution, bimodal distribution, interval, and other similar modes of precisiation.\\

\textbf{Case 3}. Suppose that local dynamics of system $S$ are expressed perceptually as:   
\begin{enumerate}
 \item if customer satisfaction is low, then purchase frequency is low
 \item if customer satisfaction is high, then purchase frequency is high
 \end{enumerate}
 This example illustrates how purchase frequency depends on customer satisfaction, an economic relationship for which no explicit differential equation or physical law may exist. Nonetheless, a perception-based description provides PrINNs with valuable insights into the system, potentially leading to a model that aligns with expert opinions to a significant extent.

Indeed, such a description of the dynamics of a system is a well-known approach in the field of fuzzy set theory. In fact, the fuzzy graph precisiation, described as a collection of fuzzy if-then rules, has played a pivotal role in a number of successful applications, such as in control systems, modeling, decision-making support systems, and so on. A simple example in this regard is the following fuzzy graph, which underlies control systems theory:
 \begin{enumerate}
 \item if distance is far, then speed is high,
 \item if distance is near, then speed is low
 \end{enumerate}
For which the overall constraining relation, representing the compressed information, can be expressed as:
\begin{equation} \label{FG-mode1-5}
R: \mu_{\text{far}}(distance) \wedge \mu_{\text{high}}(speed) + \mu_{\text{near}}(distance) \wedge \mu_{\text{low}}( speed)
\end{equation}

To recapitulate, case (1) highlights the fact that in the absence of differential equations, it may still be possible to describe some of the system’s state dynamics over an approximating time, locally by words representing the perception-based rules governing the system elastically. Moreover, case (2) aims to emphasize the application of fuzzy graphs in the precisiation of perceptions regarding the variation of system states, i.e., perception-based derivatives. Furthermore, as shown in case (3), fuzzy if-then rules can be viewed as perception-based rules precisiated in the fuzzy graph mode, allowing the networks to be informed of them in the form of compressed information.  
In addition, as will be discussed in the next section, PrINNs in which perception-based rules are precisiated in the fuzzy graph mode result in fuzzy informed neural networks (FINNs), offering notable advantages, one of which is the ability to design a data-driven control system where the network does not necessarily need to be pre-trained. \\

 The discussion in this section—whether or not its governing differential equation is known—on describing a dynamical system through perception and various approaches for its precisiation, lays the groundwork for introducing PrINNs, which will be explored in the next section.

\section{Perception-Informed Neural Networks}
PrINNs can be viewed as a generalization of informed neural networks. In fact, PrINNs are neural networks informed of system dynamics, relationships between system states, or the overall impression of certain parts of the system or features that the network should be aware of, whether in a general, local, exact, or approximate manner, through perception described in words or propositions. Broadly speaking, PrINNs serve as a platform to promote insights into different informed neural networks, introduce modern data-driven models, and advance the boundaries of computational science and engineering. In a more specific sense, the function of PrINNs is primarily to enable the network to model a system (e.g., for prediction and analysis), solve a differential equation, find an IDE solution, approximate unknown parameters, or combine such functions in an imprecise environment.

\begin{figure}[ht!] 
	\centering
	\hspace*{-1cm}
	\includegraphics[scale=0.18]{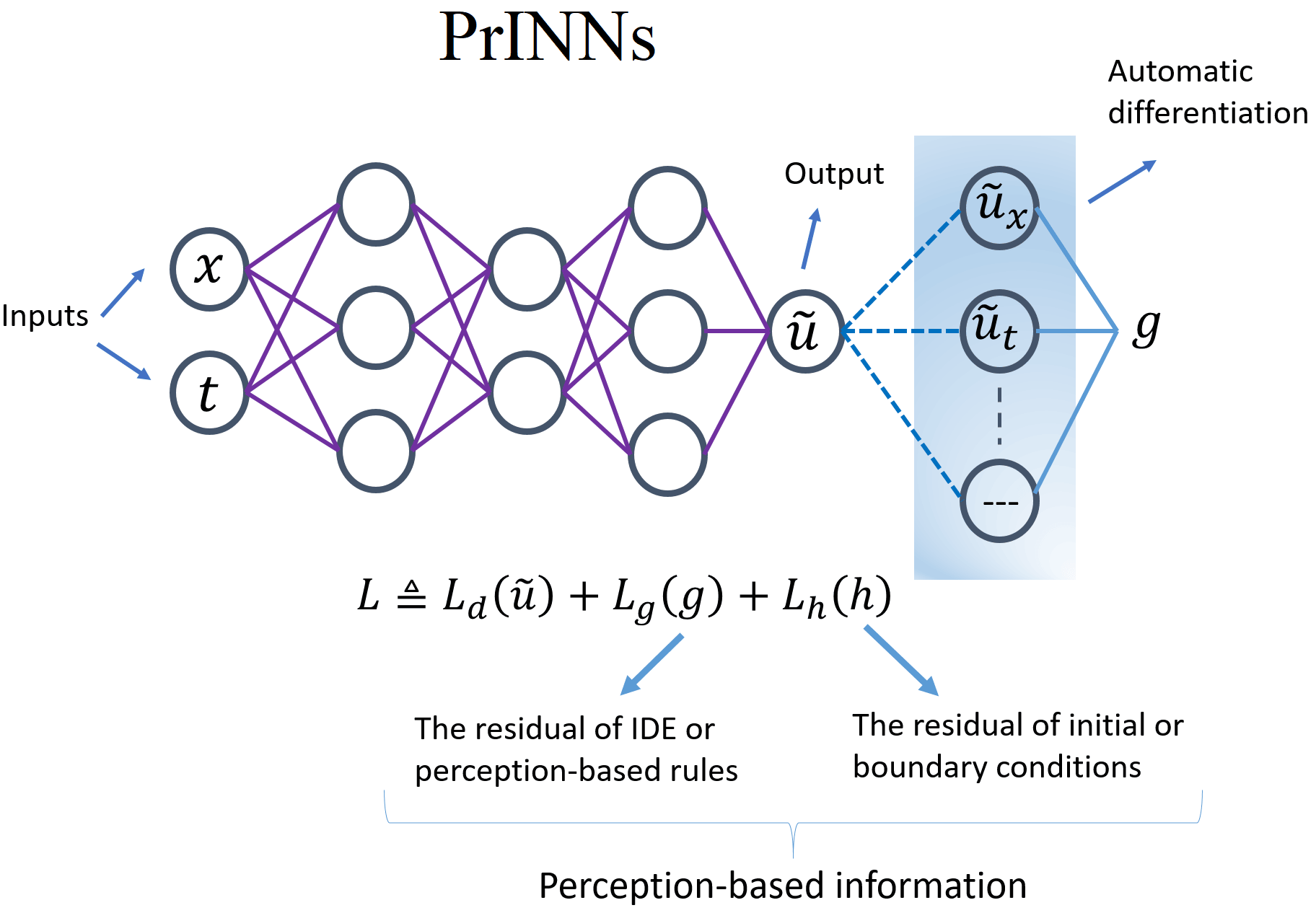}
    \caption{Perception informed neural networks.}
    \label{fig4}
\end{figure}

In PrINNs, perception, or a function of perception, should be integrated into the network's loss function. However, due to the inherent imprecision of perception, its precisiation in one mode or a mixture of modes should be considered in the loss function. Thus, PrINNs must be precisiated, and their precisiation is linked to the perception terms in the loss functions, as shown in Fig. \ref{fig4}. Simply put, for the system $ S $ with the constraint $ R $ resulting from perception, and the description of $ S $ expressed in the GC form as $ GC(S) \ isr \ R $, the general approach to setting up PrINNs is to integrate $ R $ into the loss function, with the mode of precisiation indicated.\\

The following introduces some PrINNs briefly, categorized based on the modes of precisiation and whether or not the system's differential equations are known.

\subsection{PrINNs with Known Differential Equations} 
Without losing generality and for the sake of simplicity, let us assume that the differential equation of the dynamical system $S$ is in the following form:
\begin{equation} \label{PrINNs-section1-1}
\dot{x}(t) = f(t,x;\lambda)
\end{equation}
where $ t \in [t_0, T] $, $ f $ is generally a nonlinear function, and $ \lambda $ denotes a set of parameters, e.g., $ \lambda = \{\lambda_1, \lambda_2, ..., \lambda_n\} $. Additionally, the initial condition is given by $ x(t_0) = x_0 $. Moreover, both the parameters and the initial condition are assumed to be described perceptually through words or propositions. In this context, as explained in the preceding section, the differential equation (\ref{PrINNs-section1-1}) is effectively an ordinary IDE.

To construct a PrINN for $S$, we first express the IDE (\ref{PrINNs-section1-1}) and the initial condition as residuals:

\begin{equation} \label{PrINNs-section1-2}
\begin{array}{l}
g(t, x;\lambda)=\dot{x}(t) - f(t,x;\lambda) \\
h(x(t_0)) =  x(t_0) - x_0
\end{array}
\end{equation}

Then, the imprecise functions $ g(t, x;\lambda) $ and $ h(x(t_0)) $ need to be precisiated. Let us symbolically represent the precisiation in question as $ GC(g(t, x;\lambda)) $ and $ GC(h(x(t_0))) $. Subsequently, depending on the mode of precisiation and the intended goal of PrINNs, such as modeling, finding the differential equation solution, or determining unknown parameters, a set of loss functions is established to inform the network or a collection of networks constituting the PrINNs. It should be noted that, fundamentally, it is the precisiation mode that shapes PrINNs. In the following, PrINNs in some of these modes are introduced.\\

\textbf{Singular mode}. The simplest and yet important mode is when PrINNs include IDEs governing the dynamical system that are precisiated in the singular mode. Therefore, $ g(t, x;\lambda) $ and $ h(x(t_0)) $ are crisp functions.

  Then, the loss functions for the initial condition and the differential equation residual denoted by \( \mathcal{L}_h \) and \( \mathcal{L}_g \), respectively,  are defined typically using the mean squared error\footnote{Mean squared error is one of many alternative loss functions.} as follows:

\begin{equation}
\mathcal{L}_h(h(\hat{x}(t_0))) = \left\| h(\hat{x}(t_0)) \right\|^2 = \left\| \hat{x}(t_0) - x_0 \right\|^2,
\end{equation}

\begin{equation}
\mathcal{L}_g(g(t_i, \hat{x}_i; \lambda)) = \frac{1}{N} \sum_{i=1}^{N} \left\| g(t_i, \hat{x}_i; \lambda) \right\|^2 = \frac{1}{N} \sum_{i=1}^{N} \left\| \dot{\hat{x}}_i - f(t_i, \hat{x}_i; \lambda) \right\|^2,
\end{equation}

where \( \hat{x}({t_0}) \) is the neural network prediction at the initial time \( t_0 \), and \( x_0 \) is the given initial condition, and  $ \hat{x}_i $ is the output of network, i.e. the predicted value or solution of differential equation at $ t_i $. The term \( \dot{\hat{x}}_i \) denotes the derivative of the network output at point \( t_i \), computed via automatic differentiation. Moreover, the summation is taken over a set of collocation points \( \{t_i\}_{i=1}^N \). The loss \( \mathcal{L}_h \) enforces the initial condition, while \( \mathcal{L}_g \) enforces agreement with the governing differential equation.
 
 The total loss function\footnote{In the case for which there are some boundary conditions, the total loss function includes also the loss function of boundary conditions.} is then given by: 
 
\begin{equation} \label{PrINNs-section1-3}
 \mathcal{L} \triangleq \mathcal{L}_h (h(\hat{x}(t_0))) + \mathcal{L}_g (g(t_i, \hat{x}_i;\lambda)) 
   \end{equation}  

 It should be noted that this setting is basically helpful to find the solution of (\ref{PrINNs-section1-1}). Nonetheless, if modelling or finding some unknown parameters are supposed to be obtained, then the total loss function may be considered as
\begin{equation} \label{PrINNs-section1-4}
 \mathcal{L} \triangleq \mathcal{L}_d (\hat{x}_i) + \mathcal{L}_h (h(\hat{x}(t_0))) + \mathcal{L}_g (g(t_i, \hat{x}_i;\lambda)) 
\end{equation}  

In fact, the term \( \mathcal{L}_d(\hat{x}_i) \) corresponds to the data loss, which is introduced when observational data are available. It measures the discrepancy between the network predictions and observed data at certain sampled points. The loss is defined as a mean squared error:

\begin{equation}
\mathcal{L}_d = \frac{1}{N_d} \sum_{i=1}^{N_d} \left\| \hat{x}_i - x^{\text{data}}_i \right\|^2,
\end{equation}

where \( x^{\text{data}}_i \) denotes the observed value at time \( t_i \), and \( \hat{x}_i \) is the neural network prediction at the same point. While this term is often used for parameter identification in inverse problems, it is equally valuable when the goal is to model or emulate a physical system based on known dynamics. In such cases, \( \mathcal{L}_d \) acts as a regularizing or guiding term that enforces agreement with real-world data, complementing the physical constraints imposed by the differential equation. \\

In this perspective, the resulting PrINNs coincide with the well-known PINNs. As a result, neural networks derived from PINNs, where the underlying differential equations are deterministic or crisp, can be considered as a case of PrINNs precisiated in the singular mode. Examples in this regard include physics-informed Kolmogorov-Arnold neural networks (PIKANs) \cite{PIKANs}, separable physics-informed Kolmogorov-Arnold networks (SPIKANs) \cite{SPIKANs}, physics-informed Gaussians (PIGs) \cite{PIGs}, Multi-Level physics-informed neural networks (ml-PINNs) \cite{ml-PINNs}, generalized Monte Carlo PINNs (GMC-PINNs) \cite{GMC-PINNs}, Monte Carlo fractional PINNs (MC-fPINNs) \cite{MC-fPINNs}, and so forth. Thus, symbolically, we may write:

 $$ GC(\text{PrINNs})\big|_{r=s} = \text{PINNs}, \text{PIKANs}, \text{SPIKANs}, \text{PIGs}, \text{ml-PINNs}, \text{GMC-PINNs}, \text{MC-fPINNs} $$\\

\textbf{Possibility Distribution Mode}. This mode is associated with fuzzy mathematics and fuzzy calculus. Thus, PrINNs precisiated in the mode of possibility distribution include fuzzy sets, fuzzy functions, or fuzzy differential equations. More concretely, in the possibility distribution mode, $ g(t, x;\lambda) $ and $ h(x(t_0)) $ are residuals of the fuzzy differential equation and fuzzy initial condition corresponding to IDE (\ref{PrINNs-section1-1}). To incorporate the fuzzy residuals,  $g$ and $h$, into the loss functions, a suggested approach is to employ the horizontal membership functions (HMFs) \cite{HMF-1, HMF-2, gr} of  $g$ and $h$, denoted by $\mathcal{H}(g)$ and $\mathcal{H}(h)$. Using HMFs of fuzzy numbers and fuzzy functions is a well-known approach in fuzzy mathematics and fuzzy calculus, as detailed in \cite{gr-1, gr-2, gr-3}. In fact, the use of the HMF of a fuzzy number, or fuzzy function, makes it possible to access the granules of the fuzzy number or fuzzy function, which are crisp values or functions of the membership degree, $\mu$, and the relative-distance-measure (RDM) variable, denoted by $ \alpha $. Let us suppose that the granules of $g$ and $h$ obtained by their HMFs are as follows:

\begin{equation} \label{PrINNs-section1-5}
\begin{array}{l}
 \mathcal{H}(g) \triangleq g^{gr}(t, x,\mu,\alpha_{\lambda}) \\
  \mathcal{H}(h) \triangleq h^{gr}(x(t_0), \mu,\alpha_{x_0})
\end{array}
\end{equation}
where $ \alpha_{\lambda} $ and $ \alpha_{x_0} $ are the RDM variables of $ \lambda $ and $ x_0 $ precisted in the possibility mode and they have been assumed that their precisiation has been resulted in fuzzy numbers. Thus, in this setting, the loss functions $ \mathcal{L}_g $ and $ \mathcal{L}_h $ are defined as follows
\begin{equation} \label{PrINNs-section1-6}
\begin{array}{l}
 \mathcal{L}_g(\mathcal{H}(g)) \triangleq \mathcal{L}_g (g^{gr}(t_i, \hat{x}_i,\mu,\alpha_{\lambda})) \\
  \mathcal{L}_h(\mathcal{H}(h)) \triangleq \mathcal{L}_h (h^{gr}(\hat{x}(t_0), \mu,\alpha_{x_0}))
\end{array}
\end{equation}

With the understanding that $ \mu,\alpha_{\lambda} $ and $ \alpha_{x_0} $ are learnable parameters, we proceed to define the total loss function. An additional factor representing the possibility degree or feasibility degree should be considered. This factor should inform PrINNs that, although moving away from the most probable values corresponding to $ \mu=1 $ is feasible, the degree to which our perception of $ \lambda $ and $ x_0 $ is satisfied by the granules of $ \lambda $ and $ x_0 $, or more generally, the granule of the fuzzy differential equation solution, decreases. In other words, $ \mu $ may be viewed as the degree of perception satisfaction of the granule values representing the imprecision in the  propositions.
Therefore, a decreasing function $ f $, mapping $ [0, 1] $ to a subset of the real numbers set $ \mathcal{R} $, with $ f(1)=1 $, may be considered as the possibility factor. A candidate for the possibility factor may take the form of $ M^{1-\mu} $, where $ M>1 $. Thus, the total loss function may be defined as:

\begin{equation} \label{PrINNs-section1-7}
 \mathcal{L} \triangleq \mathcal{L}_d (\hat{x}(t_i)) + M^{1-\mu}\mathcal{L}_h(\mathcal{H}(h)) + M^{1-\mu}\mathcal{L}_g(\mathcal{H}(g))
\end{equation}
It should be noted that the possibility factor should be considered for the fuzzy term granules in the total loss function.\\ 

What has been explained to incorporate fuzzy differential equations into PrINNs, as shown in (\ref{PrINNs-section1-7}), is associated with the capability of PrINNs when utilized for modeling or finding unknown parameters in an uncertain environment. This configuration of PrINNs is referred to as fuzzy calculus-informed neural networks (FcINNs).

\begin{note}
In the FcINNs explained above we have been applied granular calculus approach. Nonetheless, in the area of fuzzy calculus and mathematics, there are also some other approaches which can be utilized in FcINNs among which are the concepts related to generalized Hukuhara differentiability, the calculus for linearly correlated fuzzy functions\cite{afclin-2}, and fuzzy interactive calculus\cite{afclin-1}. 
\end{note}

 However, if the goal is to obtain the solution of FDEs, then various approaches may be suggested, one of which has been introduced in \cite{fPINN}.

Briefly, to find the solution of an FDE through PrINNs, two networks are designed: one acts as an approximator for the minimum and maximum possible primary outputs of the FDE solution, while the other determines the corresponding input fields that lead to these extreme solutions. This configuration of PrINNs is called fuzzy physics-informed neural networks (fPINNs), where the interval physics-informed neural network (iPINN) serves as a building block. The fPINN framework essentially consists of multiple iPINNs, each solving for different $\mu$-level sets of the fuzzy input fields, allowing for a possibilistic characterization of uncertainty in fuzzy differential equations.
In fact, what has been done in studies such as \cite{fPINN} is also the precisiation of PrINNs in the mode of possibility distribution. As such, we can symbolically conclude that:

 $$ GC(\text{PrINNs})\big|_{r=f} = \text{FcINNs}, \text{fPINNs} $$
 

\textbf{Probability Distribution Mode}. PrINNs in the probability distribution mode deal with either stochastic or random processes. Here, the focus is on the precisiation of PrINNs that ultimately leads to results that themselves become a random field. Simply put, in the probability distribution mode, the residuals $ g(t, x;\lambda) $ and $ h(x(t_0)) $, corresponding to IDE (\ref{PrINNs-section1-1}), include random parameters. In this context, one notable approach is the PINN under uncertainty (PINN-UU) \cite{PINN-UU}, proposed to solve differential equations with random physical parameters. 

To briefly explain PINN-UU from the perspective of PrINNs: in this framework, by considering IDE (\ref{PrINNs-section1-1}), the network takes $(t,\lambda)$ as input, where $\lambda$ represents parameters precisiated in the probability distribution mode. The network is trained to satisfy the differential equation by drawing many Monte Carlo samples of $\lambda$. The resulting trained network, once converged, provides a surrogate solution that can be evaluated at any new random draw of $\lambda$ efficiently.
 To illustrate further, let us consider the isotherm one-dimensional advection-dispersion equation coupled with Langmuir sorption as an example, adopted from \cite{PINN-UU}, which can be expressed as the following IDE:
\begin{equation}\label{PrINNs-section1-8}
\begin{cases}
    u_t \left[ 1 + \frac{\rho_b}{\theta} \left( \frac{K_l Q}{1 + K_l u^2} \right) \right] - \left( D u_{xx} - v u_x \right) = 0, \\
    v u - D u_x = Q_{\text{in}},  \\
    u_x = g_N,  \\
    u = u_0, 
\end{cases}
\end{equation}

where $ D $ is about 18, $ K_l $ is approximately 0.65, and $ Q $ is around 5. Then, by precisiating $ \lambda =\{D, K_l, Q\}  $ in the probability distribution mode, the PrINN reduces to PINN-UU. In addition, in the probability distribution mode, PrINNs can be precisiated to networks integrating stochastic differential equations. A case in point is Bayesian physics-informed neural networks (B-PINNs) \cite{B-PINNs}, in which the proposed approach, based on Bayesian techniques, aims to find the solution or determine unknown parameters of partial differential equations with random parameters in a noisy environment. Other examples include stochastic physics-informed neural ordinary differential equations (SPINODE) \cite{SPINODE}, and statistics-informed neural networks (SINN) \cite{SINN}. As a result, symbolically, we can write:

$$ GC(\text{PrINNs})\big|_{r=p} = \text{PINNs-UU}, \text{B-PINNs}, \text{SPINODE}, \text{SINN} $$

This means that neural networks integrating random or stochastic differential equations governing dynamical systems, as a means to capture quantified uncertainty and improve system predictions, underlie PrINNs, provided that the precisiation mode is probability distribution. \\

\textbf{Mixture of Experts}. MOE is one of the main contributions of PrINNs which provides the way to encode different perception precisiations of experts in the network. Networks underlying PrINNs through MOE are called  mixture of experts informed neural networks (MOEINNs). In MOEINNs, the perception regarding the parameters, initial or boundary conditions pertaining to the physics laws or differential equations governing the dynamical system is precisiated in a mixture of modes. In fact, MOEINNs with the capability of embracing a mixture of experts knowledge come with some remarkable advantages, driving advancements in scientific machine learning and computational science and engineering. What follows sheds a light on some of such advantages. \\

\textbf{1. Discovering New Differential Equations}. 
As mentioned in the preceding section, IDEs can be considered the foundation of differential equations, with the understanding that other differential equations can be viewed as a mode, or a mixture of modes, of the precisiation of IDEs. In fact, what has been considered so far in informed neural networks, or informed machine learning, has been based on incorporating differential equations derived from the precisiation of IDEs in a single mode. Examples in this regard include PINNs, fPINNs, B-PINNs, and iPINNs, which are the result of precisiation in the singular mode, i.e., r=$ s $, possibility distribution mode, i.e., r=$ f $, probability distribution mode, i.e., r=$ p $, and interval mode, i.e., r=$ cg $, in that order.
In an \textit{inverse problem}, the effort to identify the differential equation by characterizing its unknown parameters through the network has been based on a single mode of precisiation of IDEs. In other words, and more precisely, in inverse problems, the unknown parameters have been subject to a \textit{homogeneous precisiation}. Examples include deterministic parameters in crisp differential equations pertaining to the singular mode of precisiation, random parameters in random differential equations pertaining to the probability distribution mode of precisiation, and interval parameters in interval differential equations pertaining to the interval mode of precisiation, and so forth. Nonetheless, parameters can be subject to \textit{non-homogeneous precisiation} through MOEINNs.

As a simple example, let us assume that $ \lambda = \{D, K_l, Q\} $ in (\ref{PrINNs-section1-8}) is unknown, and its non-homogeneous precisiation is such that $D$ is precisiated in the mode of probability distribution and characterized by the normal distribution function $ D = \mathcal{N}(18,1) $; $K_l$ is precisiated through the possibility distribution mode by a triangular fuzzy number $ K_l = (0.05, 0.65, 0.1) $, and $Q$ is precisiated in the mode of bimodal distribution as follows:
\begin{equation}
Q \ \text{isbm} \ p_1\,/\!\,\textit{less than 5} + p_2\,/\!\,\textit{approximately 5} + p_3\,/\!\,\textit{near 10}
\end{equation}
where expressions such as $p_1\,/\!\,\textit{less than 5}$ indicate that the probability that $Q$ is less than 5 is $ p_1 $. In addition,  \textit{less than 5}, \textit{approximately 5}, and \textit{near 10} are unknown fuzzy numbers associated with the corresponding unknown probabilities $p_1, p_2,$ and $ p_3 $,  respectively. 
Thus, from this perspective, the IDE shown in (\ref{PrINNs-section1-8}), with the above non-homogeneous precisiation, gives rise to a new type of differential equation that has not been previously studied, but may be effectively addressed and identified using MOEINNs. As a result, the utilization of MOEINNs provides a principled foundation for discovering previously unknown forms of differential equations.

\begin{note}
It should be emphasized that the representation of IDE (\ref{PrINNs-section1-8}) using the aforementioned non-homogeneous precisiation is intended purely as an illustrative example. Its purpose is to demonstrate the conceptual potential of MOEINNs in handling heterogeneous forms of information, and should not be interpreted as the definitive or exclusive formulation of such a differential equation.
\end{note}

\textbf{2. Expanding the Space of Model Discovery.} By incorporating a mixture of modes of precisiation, MOEINNs are capable of encoding the knowledge of diverse experts, as well as valuable experimental information related to the parameters involved in IDEs. This flexible representation significantly expands the space of candidate models, potentially leading to better assimilation of observed data and, ultimately, more accurate predictions and more insightful analysis.\\

\textbf{3. Transformed-Knowledge Informed Neural Networks}. Considering a mixture of expert knowledge can lead to new notions, criteria, or meta-information by which the network's performance and prediction capabilities may be significantly improved. Neural networks that encode a transformation of diverse knowledge into the network's structure may be called transformed-knowledge informed neural networks (TKINNs). In fact, there have been several studies aimed at improving networks' ability to capture information associated with the underlying physics of the problem more efficiently through various approaches, such as input or output transformation \cite{ref29, ref31}, augmentation \cite{ref30}, and network architecture modification \cite{ref32}, which are primarily concerned with feature extraction and expansion.
Nevertheless, the thesis of TKINNs is to employ meta-information extracted from the interaction of specialists' knowledge to gain an enhanced insight into the dynamical system parameters, states, or to reach a consensus on specialists' viewpoints. In essence, unlike approaches that suggest the extraction or expansion of features through a transformation of information, the principle of TKINNs is based on the extraction or expansion of meta-information through a transformation of knowledge. To illustrate this, a type of TKINN is presented briefly in the following section.

Suppose a specialist precisiates the perception-based parameters of a differential equation governing a dynamical system in the probability distribution mode, while another specialist does so in the possibility distribution mode. In this case, the experts' knowledge can be handled through a MOEINN by integrating their corresponding residuals into the total loss function. However, the MOEINN lacks deep insight into the interaction of the specialists' knowledge, particularly with regard to the consistency principle between possibility and probability.
In fact, according to this principle, if an event is impossible, it must also be improbable. The concept of sureness arises from a transformation of knowledge where two different modes of precisiation, possibility and probability, are combined. This transformation yields a measure that reflects a consensus between two perspectives on precisiation and encodes consistency with the logical principle that impossibility implies improbability. By defining sureness as the product of possibility and probability \cite{sureness-paper}, a TKINN can be set up by incorporating a sureness loss function, exemplified in the following form:

\begin{equation}
\mathcal{L}_s \triangleq (1-sureness)^2 
\end{equation}
As an illustration, consider a simple example. Suppose the differential equation governing a damped harmonic oscillator, in the form of an IDE, is as follows:
\begin{equation} \label{PrINNs-section1-oscillator-1}
\left\{
\begin{array}{l}
\ddot{x}(t) + 2 \zeta \omega_n \dot{x}(t) + \omega_n^2 x(t) = 0,\\
\text{The damping ratio is approximately } 0.2, \\
x(t_0) = x_0,\\
\dot{x}(t_0) = \dot{x}_0, 
\end{array}
\right.
\end{equation}
where $ \omega_n, x_0 $ and $ \dot{x}_0 $ are deterministic. Then, the residuals corresponding to (\ref{PrINNs-section1-oscillator-1}) are as
\begin{equation} \label{PrINNs-section1-oscillator-2}
\left\{
\begin{array}{l}
g(\ddot{x}(t),\dot{x}(t), x(t); \zeta) = \ddot{x}(t) + 2 \zeta \omega_n \dot{x}(t) + \omega_n^2 x(t),\\
h_1(x(t_0)) = x(t_0) - x_0\\
h_2(\dot{x}(t_0)) = \dot{x}(t_0) - \dot{x}_0 
\end{array}
\right.
\end{equation}
In the residual $ g $ the value of parameter $\zeta $ has been described perceptually by two specialists as \textit{approximately 0.2}. Suppose one of the specialists precisiates it in the probability distribution mode with the normal distribution $ \mathcal{N}(0.2, 0.01) $, and the other specialist does so in the possibility distribution mode with a triangular fuzzy number $ (0.15,0.2,0.25) $. Then, an approach to set up the total loss function of a TKINN may be as follows:
\begin{equation} \label{PrINNs-section1-oscillator-3}
 \mathcal{L} \triangleq \mathcal{L}_d (\hat{x}_i) + \mathcal{L}_{h_1} (h_1(\hat{x}(t_0))) + \mathcal{L}_{h_2} (h_2(\dot{\hat{x}}(t_0))) + M^{1-\mu}\mathcal{L}_g(\mathcal{H}(g)) + \mathcal{L}_s
\end{equation}
where $ \mathcal{L}_d$ is the loss function pertaining to the measured data, and $ \mathcal{L}_{h_2}, \mathcal{L}_{h_1}$​​ are the loss functions pertaining to the initial conditions. Moreover, $  M^{1-\mu} $  is the possibility factor associated with the loss function of the fuzzy residual $ g $, which corresponds to the possibility distribution mode of the precisiation of IDE (\ref{PrINNs-section1-oscillator-1}). In addition, $ \mathcal{H}(g) $ is the HMF of $ g $, which can be characterized as:
\begin{equation} \label{PrINNs-section1-oscillator-4}
\mathcal{H}(g) \triangleq \ddot{\hat{x}}_i + 2 \mathcal{H}({\zeta}) \omega_n \dot{\hat{x}}_i + \omega_n^2 \hat{x}_i
\end{equation}
where $ \mathcal{H}({\zeta})\triangleq 0.2+(1-\mu)(0.1\alpha-0.05) $ with this understanding that $ \mu \in [0, 1] $ and $ \alpha \in [0, 1] $ are learnable parameters. In addition, $ \mathcal{L}_s \triangleq (1-sureness)^2  $ and the sureness is defined as the multiplication of probability and possibility degrees coming from of $ g $. In more detail, the probability distribution function of $ g $ denoted by $ \mathcal{N}_g  $ can be characterized as 
\begin{equation}
\mathcal{N}_g (m, \sigma^2) = \mathcal{N}(\ddot{\hat{x}}_i + \omega_n^2 \hat{x}_i + 2\omega_n \dot{\hat{x}}_i m,\ {(2\omega_n \dot{\hat{x}}_i)}^2\sigma^2)
\end{equation}
where $ m=0.2 $ and $ \sigma^2 = 0.01 $ according to the specialist knowledge stated above. In addition, $ \mathcal{N}_g $​ is a function from $ g(\ddot{x}_i,\dot{x}_i, x_i) $ to $ [0, 1] $. In other words, for each value of $ g $ at $ \ddot{x}_i,\dot{x}_i$ and $ x_i $​, the function $ \mathcal{N}_g $​ outputs the likelihood, denoted by $ \mathcal{N}_{g_i} $​​. Since $ \ddot{x}_i$ and $ \dot{x}_i$​ can be determined by automatic differentiation of the neural network output $ x_i $​ in each epoch, $ \mathcal{N}_{g_i} $​​ is also known in each epoch. Furthermore, the possibility degree of $ g $ is $ \mu $, a learnable parameter that is known in each epoch. Thus, the sureness in $ \mathcal{L}_s $ can be defined as the multiplication of $ \mu $ and $ \mathcal{N}_{g_i} $
\begin{equation}
sureness = \mu  \times \mathcal{N}_{g_i}
\end{equation}
\begin{note}
In (\ref{PrINNs-section1-oscillator-4}), although $ \ddot{\hat{x}}_i $ and $ \dot{\hat{x}}_i $ are the automatic differentiation of $ \hat{x}_i $, they are theoretically representatives of the granular derivatives \cite{gr} of $ x(t) $ denoted by $ \ddot{x}^{gr}(t, \mu, \alpha) $ and $ \dot{x}^{gr}(t, \mu, \alpha) $ in turn.  
\end{note}

The TKINN explained above, which employs sureness as the meta-information, is called sureness-informed neural networks (SINNets). Regarding TKINNs, there are two comments as follows: \\

First, TKINNs can also be derived from the precisiation in a mixture of modes, by which the IDE is precisiated as a ZDE, a bimodal differential equation that includes both an RDE and an FDE. Additionally, the notion of sureness in terms of ZDE is practically the same as what has already been explained.\\

Second, apart from sureness, there may be several other concepts or pieces of meta-information that can be employed by TKINNs, including sureness time (or acceptable time) \cite{ZDE} and compatibility degree of precisiations \cite{IDE}. In brief, sureness time guides TKINNs to become models where the prediction over a period of time exceeds a specific level of sureness, thereby assuring us that the time period is acceptable because the model’s prediction remains above a specific level. Furthermore, by utilizing the Compatibility of Precisiations (CoP) principle, TKINNs can be empowered to integrate a transformation of diverse homogeneous precisiations, explaining how compatible the experts' precisiations are with the precisiated differential equations by which the network is informed. Simply put, TKINNs based on the compatibility degree of precisiations result in models that assure us they not only satisfy the precisiated differential equations but also the experts' knowledge about the model output, exceeding a specific level determined by the compatibility degree.

\subsection{PrINNs with unknown differential equations} 
Apart from several contributions, the main contribution of PrINNs manifests itself when they are used to create computational models of perception, offering a pathway to transfer knowledge structures to networks, or generally to machines, to improve their comprehension of complex information through perception-based rules, particularly when no known physics laws are available.
As discussed in subsection \ref{Section2-Subsection-2}, a system or parts of it may be described, either locally or generally, using perception-based rules, especially when there is no known differential equation associated with the system. In this context, PrINNs provide a way to transfer knowledge to the network, enabling it to understand the system's dynamics more effectively through perception-based rules. \\

In the fuzzy graph mode, perception-based rules governing the system typically appear in the form of fuzzy if-then rules. PrINNs in which the perception-based rules are precisiated in the fuzzy graph mode are called fuzzy informed neural networks (FINNs). To set up FINNs, a general approach is suggested as follows: For the system SS described in the GC form as $ GC(S) \  \text{isfg} \ R $ , with $ R = \sum_i^n A_i\times B_i  $, FINNs are set up by including the overall constraint $ R $, resulting from the perception, in the loss function. This can be formulated as follows:

\begin{equation}
\mathcal{L}_{R} = M(1-R)^2
\end{equation} 
with this assumption that $ A_i $ and $ B_i $ are normal. Moreover, since $ 0\leq R \leq 1 $, the coefficient $ M\geq 1 $ may be considered as the penalty weight. What follows elaborates on FINNs for the cases presented earlier in subsection \ref{Section2-Subsection-2}. 

\textbf{Case 1.} In this case, the dynamics of the state $ x(t) $ in the system $S$ with respect to time $ t $   have been locally described as 
 \begin{enumerate}
 \item[$ R_1 $]: if $t$ is small, then $ x(t) $ is large
 \item[$ R_2 $]: if $t$ is large, then $ x(t) $ is medium
 \end{enumerate}
for which the overall restriction reads
\begin{equation} \label{PrINNs-section2-1}
R(t, x(t)) = \mu_{\text{small}}(t) \wedge \mu_{\text{large}}(x(t)) + \mu_{\text{large}}(t) \wedge \mu_{\text{medium}}(x(t))
\end{equation}
Thus, the total loss function is
\begin{equation}\label{PrINNs-section2-2}
 \mathcal{L} \triangleq \mathcal{L}_d (\hat{x}(t_i)) + \mathcal{L}_R 
\end{equation}
where $ \mathcal{L}_R = M(1-R(t_i, \hat{x}_i))^2 $, and $ \hat{x}_i $ is the output of the network. It is worth noting that $ \mathcal{L}_R $ enables the transfer of knowledge to the network, allowing it to become aware of the perception-based rules governing the dynamics of the state $ x(t) $. Additionally, the loss function $ \mathcal{L}_R $ can serve as a form of regularization, preventing non-admissible outputs in specific regions. Furthermore, $ \mathcal{L}_R $ may also play a role in reducing the detrimental effects of outliers.\\

\textbf{Case 2}. In this case, it had been assumed that there was some local knowledge about dynamical variations of the state $ x(t) $ with respect to time  expressed as \\

\textit{The variation of $x(t)$ with respect to time $ t $ is small when $ t $ is near zero, and it is   approximately 2 when $ t $ is about 10}. \\

For which the overall restriction imposed on the perception-based derivatives of $ x(t) $ was given by
 \begin{equation} \label{PrINNs-section2-3}
R(t, \dot{x}(t)) = \mu_{\text{near zero}}(t) \wedge \mu_{\text{small}}(\dot{x}(t)) + \mu_{\text{about 10}}(t) \wedge \mu_{\text{approximately 2}}(\dot{x}(t))
\end{equation}
and the total loss function of FINNs is as
\begin{equation}\label{PrINNs-section2-4}
 \mathcal{L} \triangleq \mathcal{L}_d (\hat{x}(t_i)) + \mathcal{L}_R 
\end{equation}
where $ \mathcal{L}_R = M(1-R(t_i, \dot{\hat{x}}_i))^2 $, and $ \dot{\hat{x}} $​ is the automatic differentiation of $ \hat{x}_i $. In fact, by using $ \mathcal{L}_R $​, the perception-based derivatives of the state $ x(t) $ are encoded in the network, enabling FINNs to follow an elastic slope corresponding to the local knowledge about the state dynamics. Additionally, $ \mathcal{L}_R $​ can be viewed as a regularization tool, preventing the network from producing non-realistic deviations.\\

\textbf{Case 3.} In this case, the system dynamics are described as a collection of fuzzy if-then rules, which may involve all or some of the system states. For example, let us recall the simple example presented in subsection \ref{Section2-Subsection-2}.
 \begin{enumerate}
 \item if distance is far, then speed is high,
 \item if distance is near, then speed is low
 \end{enumerate}
 To encode fuzzy if-then rules like above in FINNs, the overall restriction imposed on the joint constraint of the variables involved in fuzzy rules should be first obtained. Such a constraint is as follows
\begin{equation} \label{PrINNs-section2-5}
R(x, u) = \mu_{\text{far}}(x) \wedge \mu_{\text{high}}(u) + \mu_{\text{near}}(x) \wedge \mu_{\text{low}}( u)
\end{equation}
where $ x $ and $ u $ are variables denoting distance and speed, in turn. Then, the total loss function of FINNs is set up similar to (\ref{PrINNs-section2-4}) with $ \mathcal{L}_R = M(1-R(x_i, \hat{u}_i))^2 $ in which $ x_i $ is the network input and $ \hat{u}_i $ is the output of the network.  \\

\begin{figure}[ht!] 
	\centering
	\hspace*{-1cm}
	\includegraphics[scale=0.5]{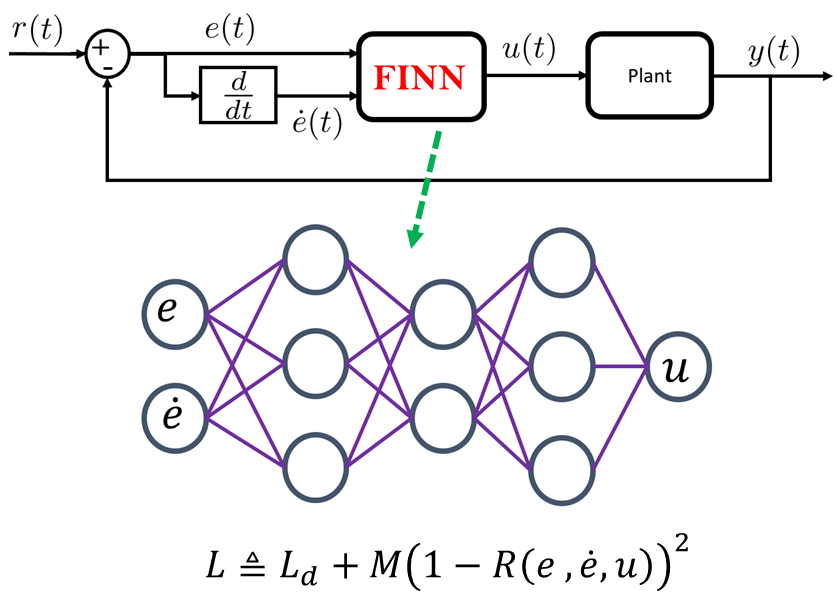}
    \caption{FINN-based controller in a closed-loop control system.}
    \label{FINNC}
\end{figure}

\begin{figure}[ht!] 
	\centering
	\hspace*{-1cm}
	\includegraphics[scale=0.4]{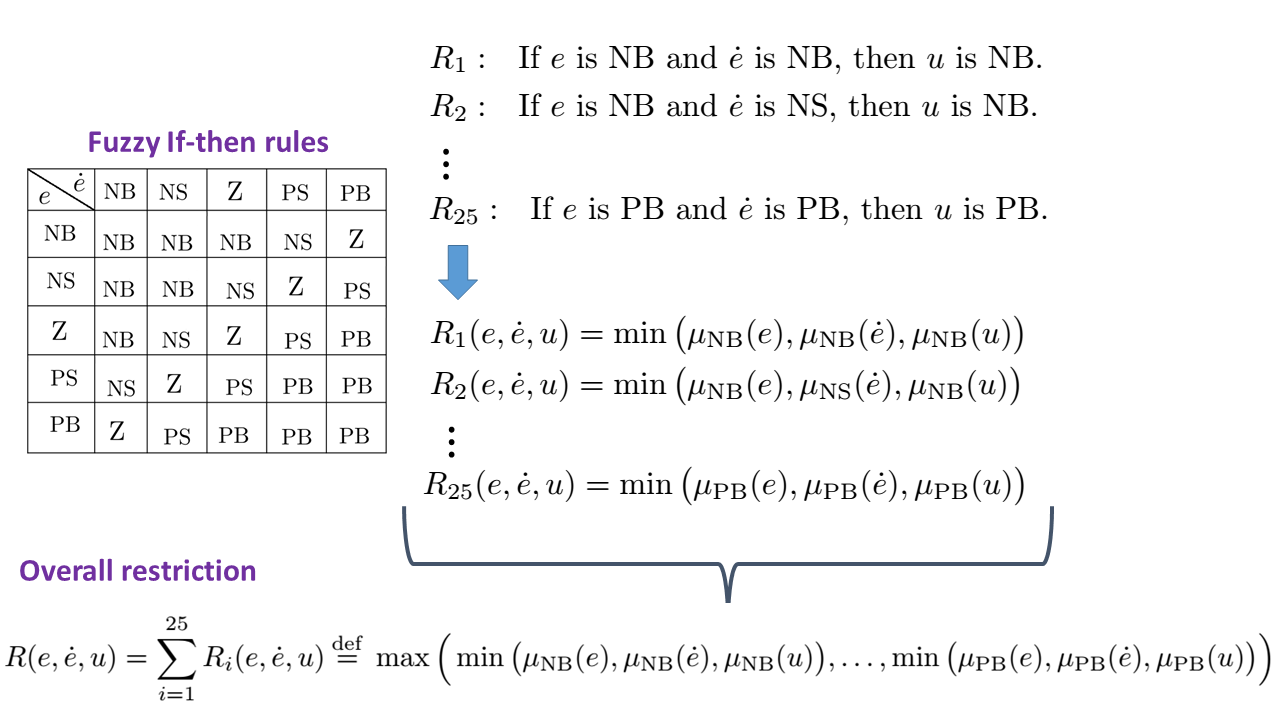}
    \caption{The overall restriction applied in the FINN-based controller.}
    \label{Rules}
\end{figure}

As a way of illustration, Fig.\ref{FINNC} shows the application of FINNs in a closed-loop control system in which the overall restriction $ R $, serving as a soft constraint on the neural network output,  comes from the knowledge of experts. In fact, the experts' knowledge demonstrate how the control signal, $u$,  should be based on the error, $e$,  and its derivative, $\dot{e}$, and such knowledge has been encoded in the form of fuzzy if-then rules, as shown in Fig.\ref{Rules}. It should be noted that in this simple and yet important example, the disjunction (or $ s $-norm) and $ t $-norm have been considered as maximum and minimum function, respectively.

Effectively, the network is encouraged to produce outputs $u$ that satisfy the control fuzzy rules, given inputs $e$ and $\dot{e}$, aligning with expert knowledge. Moreover, the closed-loop FINN-based controller functions as a real-time fuzzy controller allowing online training and inference without any need to defuzzification methods. Indeed, such a FINN-based controller can also be viewed as a data-driven controller which does not need to be trained beforehand and can be applied directly online.\\

In the cases investigated above, we have precisiated perception-based rules in the fuzzy-graph mode, yielding fuzzy if-then rules that are directly encoded into deep neural architectures, i.e. fuzzy-informed neural networks (FINNs). In essence, FINNs represent a modern class of fuzzy deep neural networks: they embed fuzzy constraints as differentiable loss terms and learn end-to-end within a unified training framework. By contrast, classical fuzzy inference systems apply their fuzzy if-then rules in separate evaluation and defuzzification steps, and adaptive neuro-fuzzy systems such as ANFIS employ a fixed, shallow layered structure with hybrid learning algorithms. It is therefore instructive to consider the advantages FINNs offer that set them apart from these established methods. What follows outlines the key differences in network architecture and flexibility by comparing FINNs, ANFIS, and traditional fuzzy inference systems.

\begin{enumerate}
  \item \textbf{Output computation}  \\
    FIS: The final output is obtained by defuzzifying the aggregated fuzzy set.  \\
    FINN: The network's output emerges naturally from its layers and learned weights, no separate defuzzification step is required.
  
  \item \textbf{Defuzzification requirement}  \\
    FIS/ANFIS: Must include a dedicated defuzzification mechanism to produce actionable numeric outputs.  \\
    FINN: Encodes the overall truth degree of the fuzzy constraints directly in the loss term, eliminating the need for any defuzzification module.
  
  \item \textbf{Knowledge- vs. data-driven orientation} \\ 
    FIS: A classic knowledge-based system, relying on expert-supplied rules and membership functions.  \\
    FINN: A data-driven framework (a subclass of PrINNs) that leverages both data and fuzzy rules in a unified training process.
  
  \item \textbf{Architectural coupling}  \\
    ANFIS: Architecturally hard-wires each FIS step into a fixed, shallow five-layer network (fuzzification, rule evaluation, normalization, defuzzification, etc.).  \\
    FINN: Decouples network depth and structure from the FIS logic, layers remain a standard deep-learning stack, and fuzzy constraints enter only through the loss.
  
  \item \textbf{Network depth and flexibility}  \\
    ANFIS: Limited by the number of FIS components (typically five layers).  \\
    FINN: Supports arbitrarily deep or wide architectures; the number of layers is not constrained by the fuzzy rule base.
  
  \item \textbf{Training paradigm}  \\
    ANFIS: Requires hybrid training (gradient descent + least squares) and often offline pre-training of rules.  \\
    FINN: Integrates fuzzy constraints into the standard end-to-end loss; can be trained (or fine-tuned) online without separate pre-training.
  
  \item \textbf{Integration into data-driven control systems}  \\
    Traditional FIS controllers require explicit plant models and defuzzification to generate control signals.  \\
    FINN: Embeds fuzzy control laws directly as loss penalties, enabling purely data-driven controller design that learns optimal actions online, no model identification or defuzzification step needed.
  
  \item \textbf{Integration with modern deep learning}  \\
    Other fuzzy-deep hybrids modify activations or insert fuzzy blocks into the network.  \\
    FINN: Leaves the internal network topology unchanged, simply adds fuzzy-rule penalties to the loss—making it easy to retrofit into any existing architecture.
  
  \item \textbf{Real-time and higher-type FIS support}  \\
    Type-2 (and higher) FIS suffer from complex, costly defuzzification at run time.  \\
    FINN: Handles type-1, type-2, or even type-n fuzzy rules by embedding them in the loss, sidestepping any defuzzification bottleneck and enabling real-time use. In other words, we can design type-1 FINNs, type-2 FINNs, and so on. 
  
  \item \textbf{Robustness to outliers and non-admissible outputs}  \\
    FIS: Lacks a built-in mechanism to penalize outputs outside feasible regions beyond rule definitions.  \\
    FINN: Uses its fuzzy-constraint loss as both a regularizer and an outlier-reduction tool, preventing unrealistic deviations.
  
  \item \textbf{Interpretability and meta-information}  \\
    FIS/ANFIS: Rules are explicit but detached from the training objective.  \\
    FINN: Retains interpretability of fuzzy rules (via loss-term contributions) and can incorporate meta-information (e.g., sureness time, compatibility degree) directly into training.
\end{enumerate}

\begin{figure}[ht!] 
	\centering
	\hspace*{-1cm}
	\includegraphics[scale=0.18]{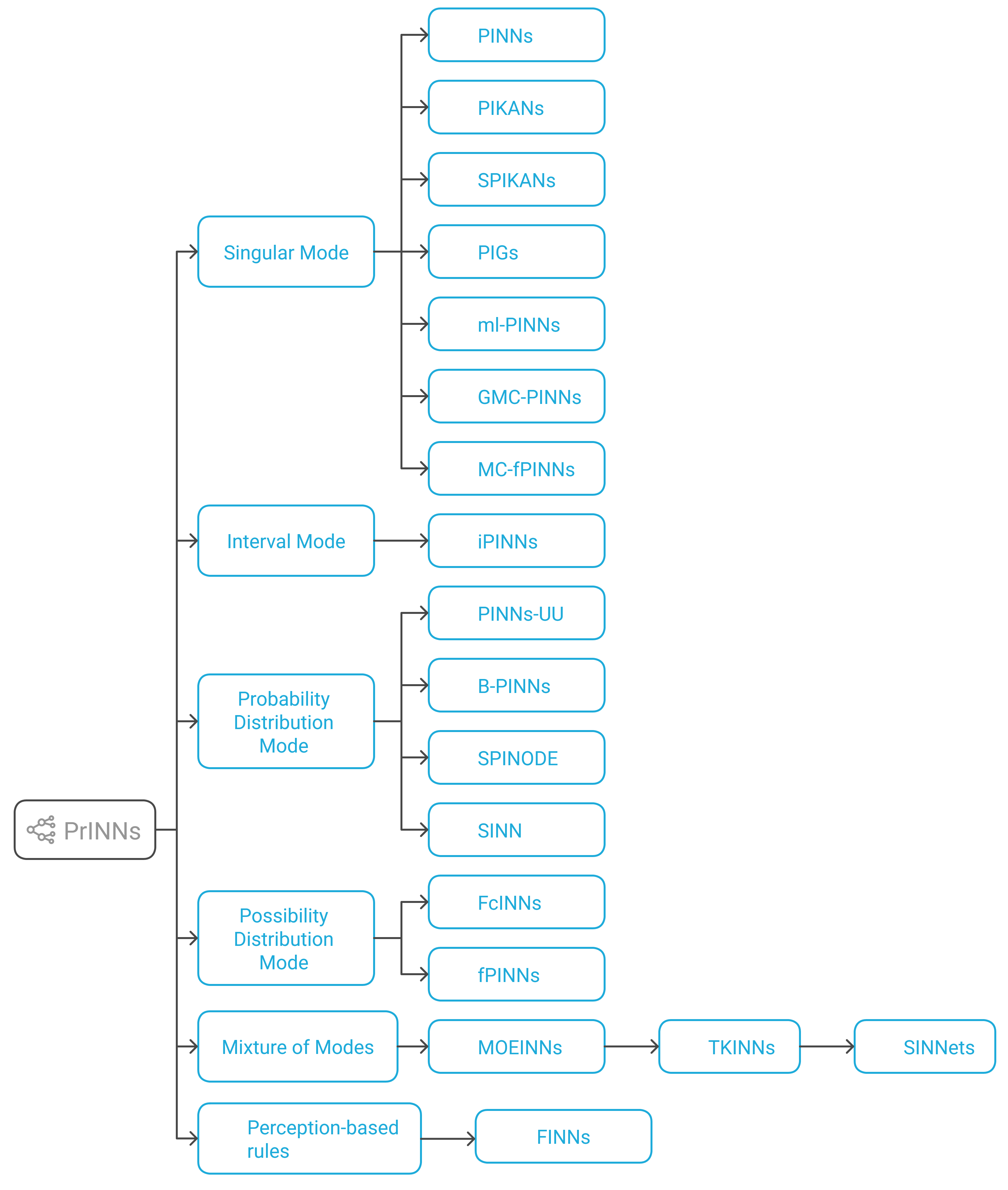}
    \caption{The variants of Perception informed neural networks.}
    \label{fig5}
\end{figure}

  To summarize, as illustrated in Fig.\ref{fig5}, PrINNs offer a distinctive platform for investigating the foundations of informed neural networks, providing valuable insights into how different types of information, ranging from deterministic laws to perceptual rules, can enrich the network's understanding of system dynamics.
  
\section{Conclusion}
PrINNs are in effect a novel framework that extends traditional neural networks by integrating perception-based information into their learning process. The main contributions of PrINNs lie in their ability to inform neural networks not only through known physics laws or differential equations but also through imprecise, perception-driven rules, allowing the networks to model complex systems even when no explicit physics laws are available. This capability makes PrINNs highly versatile, applicable to both systems with known dynamics and those where only perceptual or fuzzy-based information is accessible.\\

PrINNs provide a unique platform for exploring the origins of informed neural networks, offering insights into how various types of information, from deterministic laws to perceptual rules, can enhance network understanding of system dynamics. Key models within this framework, such as mixture of experts informed neural networks, transformed-Knowledge informed neural networks, and fuzzy-informed neural networks, introduce innovative approaches for incorporating expert knowledge and handling uncertainty in data and system dynamics.\\

PrINNs also expand the boundaries of computational science and engineering by enabling the discovery of previously unknown forms of differential equations and improving model capability through the integration of expert-driven perception. The flexibility to incorporate various modes of perception precisiation allows PrINNs to serve as a powerful tool for advancing data-driven modeling and system analysis, particularly in domains where conventional approaches may fall short. The future development of PrINNs promises to deepen our understanding of both machine learning models and the complex dynamical systems they aim to model.

\end{document}